\documentclass[a4paper,conference]{IEEEtran}

\ifCLASSINFOpdf
\else
\fi

\usepackage{mypackages}
\usepackage{mydefs}

\begin{document}
\title{CoCoLoT: Combining Complementary Trackers \\ in Long-Term Visual Tracking}

\author{
\IEEEauthorblockN{Matteo Dunnhofer and Christian Micheloni}
\IEEEauthorblockA{Machine Learning and Perception Lab\\
University of Udine\\
Udine, Italy}}

\maketitle

\begin{abstract}
How to combine the complementary capabilities of an ensemble of different algorithms has been of central interest in visual object tracking. A significant progress on such a problem has been achieved, but considering short-term tracking scenarios. Instead, long-term tracking settings have been substantially ignored by the solutions. In this paper, we explicitly consider long-term tracking scenarios and provide a framework, named CoCoLoT, that combines the characteristics of complementary visual trackers to achieve enhanced long-term tracking performance. CoCoLoT perceives whether the trackers are following the target object through an online learned deep verification model, and accordingly activates  a decision policy which selects the best performing tracker as well as it corrects the performance of the failing one. The proposed methodology is evaluated extensively and the comparison with several other solutions reveals that it competes favourably with the state-of-the-art on the most popular long-term visual tracking benchmarks.
\end{abstract}

\IEEEpeerreviewmaketitle

\section{Introduction}
Visual object tracking can be defined as the persistent recognition and localization of a target object in consecutive video frames.
It is one of the fundamental open problems in computer vision and it has many practical applications such as video surveillance \cite{micheloni2010intelligent}, behavior understanding \cite{Dunnhofer2021trek150}, robotics \cite{RE3}, and medical image analysis \cite{Dunnhofer2020MedIA}.

Depending on the behavior of the target and the dynamics of the captured scene, a visual object tracking problem can be either divided into short-term tracking or long-term tracking \cite{Lukezic2020}.
The first scenario occurs when the target never leaves completely the camera's field of view.
This is the most popular setting represented in the community's benchmark datasets \cite{OTB,VOT2019,UAV123,TC128,NfS} and subsequently the most tackled by solutions.
Successful methodologies available today to address short-term scenarios include discriminative tracking \cite{ECO,ATOM,DiMP}, siamese networks \cite{SiamFC,SiamRPNpp,SiamGAT}, deep regression trackers \cite{GOTURN,Dunnhofer2019,Dunnhofer2021ral}, and transformers \cite{TrDiMP2,TransT,Stark}.

In the setting of long-term tracking problems the assumption of the target being always visible is relaxed. In such scenarios the object disappears by leaving the field of view or by being completely occluded by another object. 
These situations require a tracker to produce not only the target's localization but also a confidence score expressing whether the object is visible or not \cite{Lukezic2020}.
In the past, long-term trackers \cite{TLD,PTAV,SPLT,LTMU} consisted in variations of an essential scheme composed of: a short-term tracking algorithm to follow the target while visible; a re-detection operation to find again the target after its reappearance; a verification module to check if the short-term tracker and re-detector have localized the object of interest.
More recently, the properties of such complex and often inefficient pipelines are starting to be achieved implicitly with compact deep neural network-based trackers. New visual trackers such as the deep discriminative tracker SuperDiMP \cite{DiMP,VOT2020} and the transformer-based solution STARK \cite{Stark} are able to match or even surpass the long-term tracking performance of previous methodologies while performing at real-time speed.
Such efficiency makes the employment of tracker fusion strategies \cite{Yoon2012,Bailer2014,Vojir2016,Dunnhofer2020accv} appealing since the processing speed of an ensemble-based solution could be reasonable. The latter approaches demonstrated how increased tracking performance can be achieved by the careful combination of the complementary capabilities of different trackers.
However, the solutions available today focused on the fusion of trackers only in short-term scenarios and, to the best of our knowledge, yet no work explored such an approach 
for %
long-term visual object tracking.

\begin{figure}[t]
\centering
  \includegraphics[width=\columnwidth]{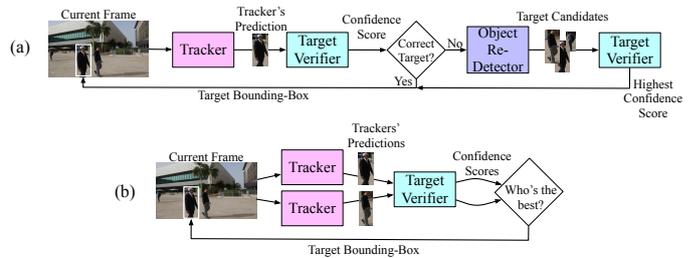}
  \caption{(a) Scheme of the procedure executed at every frame by the most popular algorithmic architecture available in the long-term visual tracking community. (b) Scheme of the procedure executed at every frame by CoCoLoT.  The proposed framework provides a simpler pipeline which just evaluates two parallel-running trackers and making them interact, ultimately removing unnecessary computationally expensive modules. } 
  \label{fig:scheme}
\end{figure}

In this paper, we try to fill such a gap by proposing a framework that aims to combine the capabilities of complementary trackers to enhance the tracking performance in long-term problems.
The framework, named CoCoLoT (\emph{Co}mbining \emph{Co}mplementary Trackers in \emph{Lo}ng-Term \emph{T}racking), is conceptually different from the approaches currently available for long-term tracking, as visualized by Figure \ref{fig:scheme}. 
Our strategy is based on a tracker evaluation function that determines if each tracker is following correctly the target. 
The function maintains an online-learned representation of the target which allows discriminating the object of interest from the background. This peculiarity is used to assess whether the localization provided by the trackers includes the appearance of the target.
The proposed evaluation strategy allows to select the best target localization among those proposed by the trackers regardless of their methodology or implementation. The outcome of the selection is exploited to make the trackers interact and correct their performance by themselves during tracking, ultimately achieving higher tracking robustness by merging their capabilities.
Overall, the main contribution of this paper is the first demonstration of tracker fusion in the context of long-term visual object tracking.
Through an extensive experimental campaign it will be demonstrated that CoCoLoT improves the performance of the underlying trackers by a good margin. In particular, we will show that the combination of two complementary tracking approaches achieves new state-of-the-art results on the most popular long-term tracking benchmarks LTB-50 \cite{Lukezic2020,VOT2019}, TLP \cite{TLP}, and LaSOT  \cite{LaSOT}.

\section{Related Work}

\subsection{Long-Term Visual Tracking}
Kalal et al. \cite{TLD} considered the long-term tracking task under the fruitful framework of tracking, learning, and detection in which:
a short-term tracker based on median-flows follows the target while visible; a learning module generates training examples during tracking for target recognition; and an online learned cascade classifier is used as target detector. 
Such a scheme has been then improved by many follow-up solutions \cite{SPLT,LTMU} which exploited deep learning models to implement the short-term tracker, the target verification module, or the detector.
Differently from such approaches, the FuCoLoT tracker \cite{Fucolot} extended the discriminative correlation filter (DCFs) approach \cite{MOSSE} to the long-term setting by optimizing multiple filters at different time scales to implement a short-term tracker and a long-term detector which predictions are then fused together. The GlobalTrack tracker \cite{GlobalTrack} made the deep siamese approach \cite{SiamFC} work in long-term scenarios by searching the target globally instead of locally in each frame.
More recently, the STARK tracker \cite{Stark} exploited transformer neural networks \cite{Attention} to implement an effective matching operation that is able to perform short-term tracking and re-detection at the same time.

Enlightened by the recent capabilities of trackers which achieve remarkable results in the long-term context without sacrifying efficiency \cite{Stark,DiMP,VOT2020}, in this work we follow a different idea with respect to the aforementioned and propose a new framework for long-term tracking aiming to combine the characteristics of complementary tracker with an effective evaluation and interaction strategy.

\subsection{Tracker Fusion Strategies}
Different approaches have been proposed to fuse the execution of multiple trackers while tracking.
Yoon et al. \cite{Yoon2012} made different trackers interact by exploiting a probabilistic approach based on particle filters.
The MEEM tracker \cite{MEEM} later provided a multi-expert framework where trackers are fused via a procedure based on entropy minimization.
Wang et al. \cite{Wang2014} and Vojir et al. \cite{Vojir2016} used variations of Hidden Markov models to implicitly correct an ensemble of interactive trackers.
Bailer et al. \cite{Bailer2014} used an optimization approach based on dynamic programming to fuse the predictions of multiple trackers without their interaction.
For an analogous setting, Dunnhofer et al. \cite{Dunnhofer2020accv} presented a tracker selection strategy based on a value function approximation learned offline via knowledge distillation and reinforcement learning (RL). Similarly, Song et al. \cite{Song2020} proposed an online selection policy optimized with hierarchical RL.

The main drawback of the solutions presented here is that they were studied for the fusion of trackers in the context of short-term tracking. In contrast, in this paper we focus on the long-term setting and, to the best of our knowledge, the proposed study is new in such a context.

\section{Methodology}
\label{sec:method}
The key idea of this paper is to develop an effective strategy to fuse the capabilities of complementary trackers in the context of long-term visual object tracking. Particularly, our goal is to implement a solution that achieves higher tracking performance in an online fashion by exploiting the characteristics of different trackers.
After the description of some preliminary concepts, in this section we will introduce the methodology to accomplish such an objective.

\subsection{Preliminaries}
We consider a video
$\video = \big\{ \frame_t \in \images \big\}_{t=0}^{T}$
as a sequence of frames $\frame_t$, where $\images =  \{0,\cdots,255\}^{w \times h \times 3}$ is the space of RGB images and $T \in \mathbb{N}$ denotes the number of frames.
 Let $\bbox_t = [x_t,y_t,w_t,h_t] \in \bboxes \subseteq \reals^4$ be the $t$-th bounding-box defining the coordinates of the top left corner, and the width and height of the rectangle containing the target.
The goal of a long-term tracker is to predict the bounding-box $\bbox_{t}$ that best fits the target and a confidence score $\confidence_t \in [0,1]$ that reports whether the target is visible in the frame, for all $\frame_t$.
We define a tracker as a function $\trackerspl : \images \xrightarrow{} \bboxes \times [0,1]$ that returns the target localization and confidence score for an input frame.
At the first frame $\frame_0$, the tracker is initialized with the ground-truth bounding-box $\bbox_0$ which outlines the target to be tracked.

\subsection{Tracker Combination Procedure}
We refer to our proposed combination algorithm as $\trackerours$. At every $t$, $\trackerours$ receives in input the frame $\frame_t$ and outputs $\bbox_t$ and $\confidence_t$.

\subsubsection{Execution of The Baseline Trackers}
\begin{figure}[t]
\centering
  \includegraphics[width=\columnwidth]{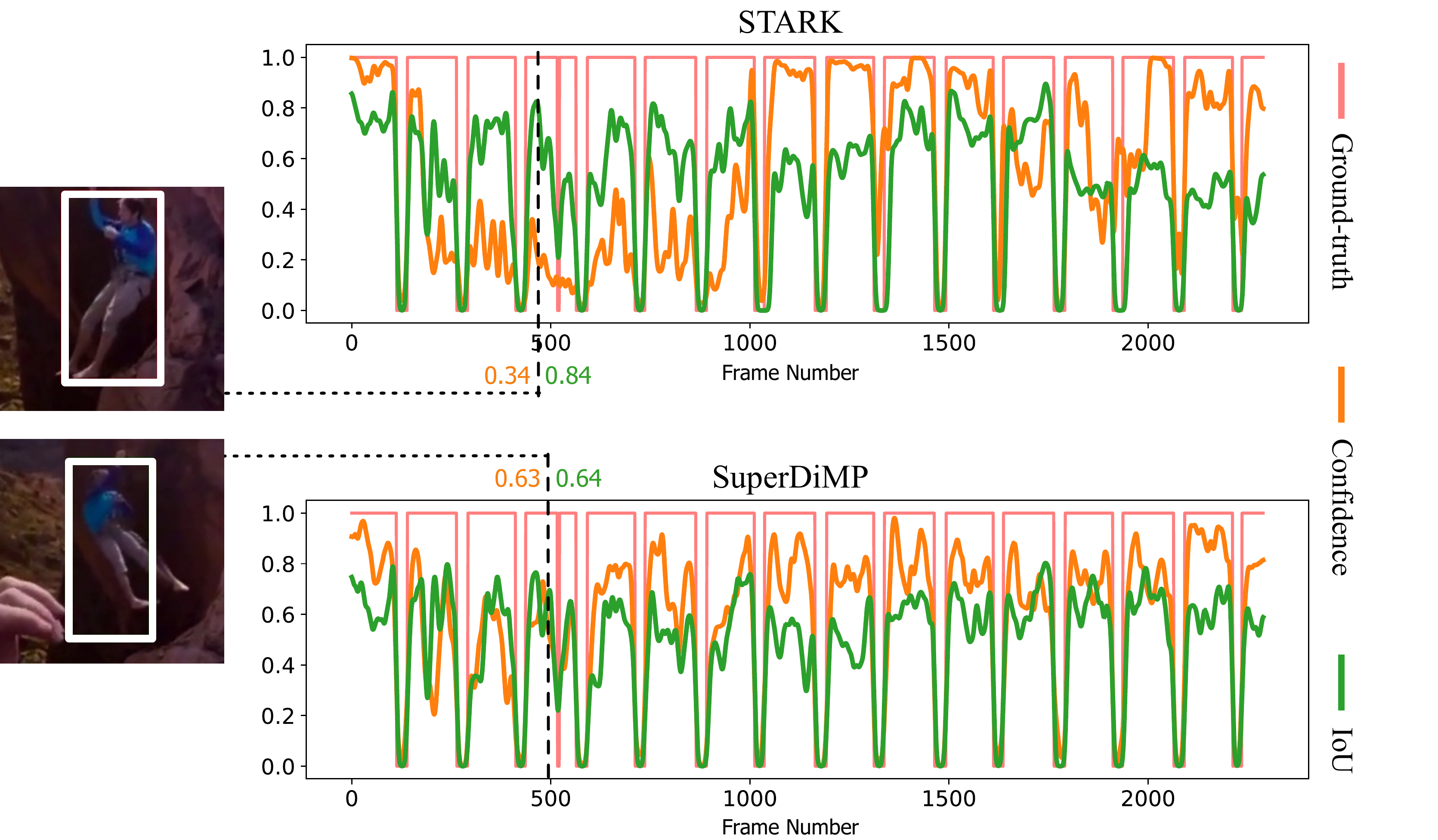}
  \caption{This figure shows the complementary characteristics of the STARK \cite{Stark} and SuperDiMP \cite{DiMP} trackers along a sequence. The first has a better ability in producing bounding-boxes better aligned with the target but that is not consistent with its ability in target presence confidence. The second provides less precise target localizations but that are more consistent with the confidence predictions.} 
  \label{fig:examples}
\end{figure}

In this work, we selected the state-of-the-art trackers STARK \cite{Stark} and SuperDiMP \cite{DiMP,VOT2020} enhanced with a meta-updater \cite{LTMU} as the baseline trackers which capabilities are fused. Such choices are motivated by the outstanding results achieved by such algorithms in the long-term setting, and because they perform tracking by different principles. 
Indeed, the first method is based on a transformer-based architecture whose tracking knowledge is acquired on a large dataset of tracking examples only through offline optimization. The second tracker instead is a deep discriminative tracker which uses an online learning mechanism to adapt a pretrained network to a new target while tracking. The aspect of performing tracking by disparate principles is especially important since complementary capabilities could benefit a combination strategy. We verified the presence of such complementary characteristics in the long-term tracking behavior of the considered trackers, and an example of outcome is proposed in Figure \ref{fig:examples}. It can be noticed that STARK manifests a better ability in producing bounding-boxes tightly fitting the targets' appearance. This behavior results in an higher Intersection-over-Union (IoU) and tells that the tracker is more spatially accurate. Such an ability however is not consistent with the confidence predictions given by the tracker which are often wrong or overconfident. On the other hand, we observe that SuperDiMP is generally less accurate in the prediction of target localization -- its IoU is lower than STARK -- but its confidence predictions are definitely more consistent with such a performance, ultimately demonstrating an increased robustness.
For a better explanation, from now on we refer to STARK as $\trackerst$ and to SuperDiMP as $\trackersd$.

In the proposed $\trackerours$ pipeline, such two baseline trackers $\trackerst, \trackersd$ are run according to their original methodology on frame $\frame_t$ when $\trackerours$ is inputted with frame $\frame_t$. By this step, they produce the respective bounding-box $\bbox_t^{(i)}$ and confidence score $\confidence_t^{(i)}$, $i=1,2$. It is worth notice that the two trackers are one independent from the other. It is hence possible to put in parallel the executions of the two in order to increase the processing speed of the combination strategy.

\subsubsection{Target Visibility Determination}
Next, $\trackerours$ determines whether $\trackerspl^{(i)}$ are correctly following the target, i.e. if it is visible in their predicted bounding-boxes.
This step is achieved by exploiting the confidence $\confidence_t^{(i)}$ which represent tracker-specific probability estimates of the target being present in the frame.
However, relying solely on $\confidence_t^{(i)}$ does not enable an effective tracker selection mechanism because such estimates can be erroneous due to overconfidence or training bias. 
We hence propose to improve such target visibility scores through a target verification module that is independent from the baseline trackers. Particularly, we employ an online learned function $\verifier : \images \times \bboxes \xrightarrow{} [0,1]$ that returns a probability estimate $\verification_t$ of the target being present in the image patch extracted from the frame $\frame_t$ considering the area determined by a bounding-box $\bbox_t$. 
This operation is inspired by the target verification operation present in different long-term tracking pipelines \cite{LTMU,VOT2019,RLTDiMP}. Such a verification step is implemented as a binary classification based on a deep neural network learned to distinguish between patches containing the target object and patches without. The architecture and learning procedure is akin to \cite{MDNet}. In short, the network is first trained offline to acquire general patch separation knowledge.
During tracking, the pretrained weights are adjusted online by an optimization procedure that uses new target appearances extracted based on the latest target localization information $\bbox_t$. A sampling procedure is performed to generate candidate target localization around $\bbox_t$. Of such samples, positive target patches are those image areas whose sampled location has an IoU greater than 0.7. Negative target patches are those resulting in an IoU of 0.3 or lower instead.
The execution of this update operation is triggered by a meta updater instance \cite{LTMU}.

Hence, in $\trackerours$, the verifier $\verifier(\cdot)$ takes the frame $\frame_t$ and bounding-box $\bbox_t^{(i)}$, and returns a tracker-independent evaluation score $\verification_t^{(i)}$ for each tracker.
Such a value is combined with the tracker's confidence as
\begin{align}
\widehat{\confidence}_t^{(i)} = \frac{\confidence_t^{(i)}+\verification_t^{(i)}}{2}.
\end{align}
$\widehat{\confidence}_t^{(i)}$ represents a more consistent target visibility estimation. We binarize such values with a 0.5 threshold to determine the status $\widehat{\presence}_t^{(i)} \in \{0,1\}$ of target visual presence in the single frames.
However, we experienced that $\widehat{\presence}_t^{(i)}$ could be wrong since the estimate is mostly based on the single-frame appearance of targets.
Given that target disappearances and reappearances are dynamic processes evolving over multiple frames we consider the $\widehat{\presence}_t^{(i)}$ present in the last $\widehat{T}$ frames to determine the actual target presence. Particularly, we say that the target is visible inside $\bbox_t^{(i)}$, and set $\presence_t^{(i)} = 1$, if
\begin{align}
\label{eq:visib}
\sum_{j=0}^{\widehat{T}-1} \widehat{\presence}_{t-j}^{(i)} > \lfloor0.75 \cdot \widehat{T} \rfloor.
\end{align}
Otherwise, we set $\presence_t^{(i)} = 0$ and the target is considered not visible in the tracker's prediction. We experimentally found $ \widehat{T} = 5$ to be a good representation of the duration of the target disappearance/appearance process.
The value $\presence_t^{(i)}$ is also used by $\trackerours$ as confidence prediction , i.e. $\confidence_t = \presence_t^{(i)}$.

\subsubsection{Target Localization Determination}
The values $\presence_t^{(i)}$ determine which tracker is currently following the target. Such information is used by $\trackerours$ to select which target bounding-box to output for frame $\frame_t$. 
If $\presence_t^{(1)}$ and $\presence_t^{(2)}$ are equal and greater than 0 $\trackerours$ selects the box produced by $\trackerspl^{(1)}$ as output $\bbox_t$ since it produces more accurate bounding-boxes in general. 
If only one of the two $\presence_t^{(i)}$ values is equal to 1 then the tracker's box corresponding to $i$ is determined and used as $\trackerours$'s output.
If both $\presence_t^{(i)}$ are zero, the bounding-box result of $\trackerspl^{(1)}$ is selected because of its better re-detection capabilities.

\subsubsection{Tracker Correction}
The predicted $\bbox_t$ is an useful resource if properly aligned on the target. We hence exploit it to correct the performance of the worse tracker. At the next $\frame_{t+1}$, $\trackerspl^{(1)}$ and $\trackerspl^{(2)}$ search for the target in an image area determined by the previously known bounding-box $\bbox_{t}^{(i)}$. We propose to modify $\bbox_{t}^{(i)}$ to match $\bbox_{t}$ when $\presence_t^{(\widehat{i})} = 1$. 
This step has a correction effect on the behavior of $\trackerspl^{(i)}$. In fact, 
$\trackerspl^{(i)}$ will search for the target in a local image area whose position is better aligned with the target position in $\frame_{t+1}$.

\section{Experimental Setup}
\label{sec:expdet}

\subsection{Datasets} 
We conducted experiments on the LTB-50 \cite{Lukezic2020} benchmark.
This dataset is used in the annual VOT challenges \cite{VOT2019,VOT2020,VOT2021}.
and it is composed of 50 videos for a total of around 215K frames densely labeled with the bounding-boxes of diverse objects (people, car, motorcycles, bicycles, boat, animals, etc.). 
Each video contains circa 10 long-term target disappearances on average each lasting for circa 52 frames.
Evaluations were also performed on the TLP dataset \cite{TLP}. This benchmark is composed of 50 video sequences comprising around 676K labeled frames. The average length of the sequences in time is over 8 minutes.
We also ran experiments on the test set of the LaSOT benchmark \cite{LaSOT}. This is composed of 280 sequences with around 690K frames and a average sequence length of 2500 frames. 

\subsection{Evaluation Protocol and Measures}
For all the experiments over all the benchmarks, we run trackers according to the standard protocol \cite{Lukezic2020,OTB,VOT2019} of initializing the tracker in the first frame and then execute it on every other frame to obtain bounding-box and confidence predictions. 
We employed established metrics to quantify the performance of our proposed solution. 
For the LTB-50 benchmark, the F-score, $\text{Precision}_{LTB}$, and $\text{Recall}_{LTB}$ metrics \cite{Lukezic2020} have been used.
On the TLP dataset, we employed the Area-Under-the-Curve (AUC) of the success plot -- referred to as $\text{Success}_{TLP}$ -- in which an IoU of 1 is set for all the frames where the tracker correctly predicts the absence of the target. Similarly, we also report the $\text{Precision}_{TLP}$ which is the AUC of the precision plot in which a bounding-box distance of 0 is set for all the frames where the tracker correctly predicts the absence of the target \cite{TLP}.
For the LaSOT benchmark, we used the AUC of the success and the precision plots referred as $\text{Success}_{LaSOT}$ and $\text{Precision}_{LaSOT}$  respectively \cite{LaSOT}.

\subsection{Improvements to STARK}
\label{sec:starkimp}
We found additional improvements to the tracking strategy of the underlying trackers to benefit the performance of our overall solution $\trackerours$. In particular, we propose to control STARK's searching area factor $\safactor$ which defines the image area size in which to look for the target. We considered $\safactor = \frac{\safactor}{2}$ in all frames in which $\presence_t^{(\widehat{i})} = 1$. Given that $\presence_t$ establishes that $\trackerours$ is following the visible target, the proposed improvement forces the tracker to better focus on it, reducing the chance of confusion due to the presence of distractors.
Moreover, we found STARK to be susceptible to wrong target size estimations after the change of the dynamic template. We propose to penalize the results of STARK by setting $\confidence_t^{(1)} = 0$ if the ratio between the aspect-ratio of $\bbox_{t-1}$ and $\bbox_{t}^{(1)}$ are not consistent with the temporal coherence of motion and scale change of a target. 
Overall, as we will show later, these improvements permit to avoid wrong target image patches to pollute the training data used by $\verifier(\cdot)$ for online adaptation, ultimately making its discriminative ability more effective.

\subsection{Implementation Details}
Code to implement the method and the experiments was implemented in Python and run on a machine with an Intel Xeon E5-2690 v4 @ 2.60GHz CPU, 320 GB of RAM and an NVIDIA TITAN V GPU.
The original implementations of the STARK and SuperDiMP trackers provided by the respective authors have been used along with the pretrained models. The verifier model $\verifier(\cdot)$ has been implemented using the PyTorch version of the MDNet tracker \cite{MDNet}. CoCoLoT runs at 5 FPS on average.

\section{Results}
In this section, we provide the results of the conducted experimental campaign. We first analyze the capabilities of CoCoLoT on the LTB-50 benchmark \cite{Lukezic2020,VOT2019} under different ablative studies and in comparison with other tracker fusion strategies. We then compare our framework with many state-of-the-art solutions on the other benchmarks described in Section \ref{sec:expdet}.

\begin{table}[t]
\fontsize{7}{8}\selectfont
	\centering
	\caption{Performance achieved by CoCoLoT on the LTB-50 benchmark \cite{Lukezic2020,VOT2019} with different coonfiguration settings. The first two rows report the performance of the underlying trackers STARK \cite{Stark} and SuperDiMP \cite{DiMP,VOT2020}. Best setup, per metric, is highlighted in bold.}
	\label{tab:spl:ablation}
	\setlength\tabcolsep{.1cm}
	\begin{tabular}{l | c c c }
		\toprule
		
		Setup & F-Score & $\text{Precision}_{LTB}$ & $\text{Recall}_{LTB}$ \\
		
		\midrule
		
		1) SuperDiMP & 0.671 & 0.691 & 0.652 \\
		2) STARK & 0.696 & 0.708 & 0.685 \\
		
		\midrule
		
		3) $\trackerours$ exploiting only maximum $\verification_t^{(i)}$  & 0.706 & 0.697 & 0.715 \\
		
		4) $\trackerours$ exploiting $\widehat{\confidence}_t^{(i)}$ and no correction & 0.710 & 0.709 & 0.711 \\
		
		5) $\trackerours$ exploiting $\widehat{\confidence}_t^{(i)}$ & 0.719 & 0.724 & 0.714 \\
		
		6) $\trackerours$ exploiting $\presence_t^{(i)}$ and no correction & 0.712 & 0.714 & 0.712 \\
		
		7) $\trackerours$ exploiting $\presence_t^{(i)}$ & 0.722 & 0.728 & 0.717 \\

		8) \quad + adaptive searching area & 0.730 & 0.741 & 0.719 \\
		9) \quad + aspect-ratio correction & \textbf{0.735} & \textbf{0.741} & \textbf{0.729} \\

		\bottomrule		
\end{tabular}
\end{table}

\begin{figure}[!ht]
\centering
  \includegraphics[width=\linewidth]{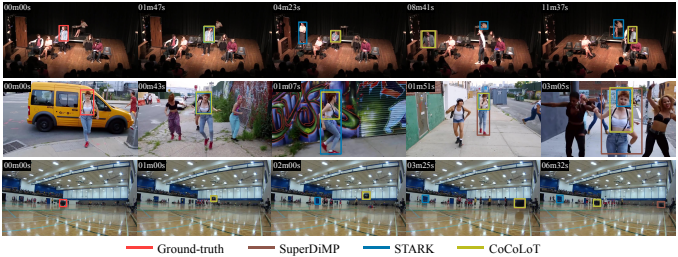}
  \caption{Qualitative examples of the tracking ability achieved by CoCoLoT in comparison with the baseline trackers STARK and SuperDiMP. The first column of images presents the first frame of each video. In the top-left corner of each frame the time elapsed since the beginning of the video is reported. Overall, our solution permits to fuse the capabilities of the underlying trackers and consequently achieve a more robust target tracking along the videos.}
  \label{fig:qualex}
\end{figure}

\begin{table}[t]
\fontsize{7}{8}\selectfont
	\centering
	\caption{Performance of CoCoLoT on the LTB-50 benchmark \cite{Lukezic2020,VOT2019} in comparison with baseline strategies that combine the capabilities of the STARK and SuperDiMP trackers (whose performances are reported in the first two rows). Best result, per metric, is highlighted in bold.}
	\label{tab:spl:baselines}
	\begin{tabular}{m{15em} | c c c }
		\toprule
		Setup & F-Score & $\text{Precision}_{LTB}$ & $\text{Recall}_{LTB}$ \\
		
		\midrule

        1) SuperDiMP & 0.671 & 0.691 & 0.652 \\
		2) STARK & 0.696 & 0.708 & 0.685 \\
		
		\midrule
		
		3) $\bbox_t^{(i)}$ and $\confidence_t^{(i)}$ average & 0.671 & 0.723 & 0.626 \\
		4) $\bbox_t^{(i)}$ and $\confidence_t^{(i)}$ average and correction of both & 0.704 & 0.711 & 0.698 \\
		5) $\bbox_t^{(i)}$ selection by maximum $\confidence_t^{(i)}$ & 0.705 & 0.703 & 0.706 \\
		6) $\bbox_t^{(i)}$ selection by maximum $\confidence_t^{(i)}$ and correction of the other & 0.675 & 0.687 & 0.675 \\

		\midrule
		
		7) TRASFUST \cite{Dunnhofer2020accv} & 0.693 & 0.701 & 0.684 \\ 
		
		\midrule
		    
		8) $\trackerours$ & \textbf{0.735} & \textbf{0.741} & \textbf{0.729} \\
        
		\bottomrule		
\end{tabular}
\end{table}

\subsection{Ablation Study}
Table \ref{tab:spl:ablation} reports the performance of $\trackerours$ on the LTB-50 benchmark \cite{Lukezic2020,VOT2019} by increasingly adding the contributions described in the Section \ref{sec:method}.
Improved performance with respect to the underlying trackers is already achieved by selecting the tracker obtaining the best $\verification_t^{(i)}$ given by the verifier (row 3). $\text{Precision}_{LTB}$ results are particularly increased by combining the trackers' confidences and the verifier scores (row 4).
Determining the target presence $\presence_t^{(i)}$ by the scores achieved in the previous $\widehat{T}$ frames additionally increases the precision (row 6). 
Row 5 and 7 show the performance is additionally improved by the interaction and correction process between trackers. 
By this setup, the performance gain with respect to the best of the underlying trackers is of around 3.7\% in F-Score, 2.8\% in $\text{Precision}_{LTB}$, and 4.7\% in $\text{Recall}_{LTB}$.
The introduction of the improvements to STARK enables the $\verifier(\cdot)$ to remove polluted samples during the training set used for online learning (row 7 and 8). This results in a more consistent optimization process that ultimately enables a better target-background discrimination.
Notice that the same strategies activated on the underlying STARK tracker based on its confidence $\confidence_t^{(1)}$ make it achieve an F-Score of 0.694 and 0.689 for the adapting searching area and aspect-ratio correction respectively.
These outcomes suggest that such strategies have to be applied carefully only when the estimation of target presence is sufficiently accurate.
Overall, the performance gain of our overall system with respect to the best of the underlying trackers is of 5.6\% in F-Score, 4.7\% in $\text{Precision}_{LTB}$, and 6.4\% in $\text{Recall}_{LTB}$.
Some qualitative examples of the performance of $\trackerours$ in comparison with the baseline trackers are presented in Figure \ref{fig:qualex}.

\subsection{Comparison with Baselines}
We compared the fusion strategy implemented by $\trackerours$ with baseline and state-of-the-art tracker fusion approaches \cite{Dunnhofer2020accv}. All the compared methods have been applied on top of the same STARK and SuperDiMP instances used in $\trackerours$. The results are given in Table \ref{tab:spl:baselines}. $\trackerours$ results much better than all the other strategies. Simply averaging the bounding-box coordinate values and respective confidence scores lowers the performance of the trackers (row 1). Correcting both trackers by their average target position and scale improves the performance of the two (row 2). Selecting the $\bbox_t^{(i)}$ for target localization based on the maximum $\confidence_t^{(i)}$ of each tracker allows to improve the performance again (row 3). But making the trackers interact in the latter setup results in a performance drop (row 4). We hypothesize this is due to the STARK's overconfidence given to bounding-box predictions having low accuracy which causes the correction of the other trackers with inaccurate boxes.
The fusion performance of TRASFUST \cite{Dunnhofer2020accv} does not allow for performance improvement of the two underlying trackers. This happens because such a fusion strategy is designed for short-term tracking settings.

\begin{table}[t]
\fontsize{7}{8}\selectfont
	\centering
	\caption{Performance achieved by CoCoLoT in the determination of the presence of the target on the LTB-50 benchmark \cite{Lukezic2020,VOT2019} in comparison with SuperDiMP and STARK. Best result, per metric, is highlighted in bold.}
	\label{tab:visibilityacc}
	\setlength\tabcolsep{.55cm}
	\begin{tabular}{l | c c c }
		\toprule
		Setup & Accuracy & Sensitivity & Specificity \\
		
		\midrule
		
		SuperDiMP & 0.828 & 0.811 & \textbf{0.907} \\
		STARK & 0.861 & 0.860 & 0.811 \\
		
		\midrule
		    
		$\trackerours$ & \textbf{0.925} & \textbf{0.937} & 0.782 \\
        
		\bottomrule		
\end{tabular}
\end{table}

\begin{table}[t]
\fontsize{7}{8}\selectfont
	\centering
	\caption{Performance achieved by $\trackerours$ on the LTB-50 benchmark \cite{Lukezic2020,VOT2019} while considering different number of frames $\widehat{T}$ to determine the visibility of the target. Best result, per metric, is highlighted in bold.}
	\label{tab:spl:visibilityframes}
	\setlength\tabcolsep{.37cm}
	\begin{tabular}{l | c c c c c}
		\toprule
		\# Frames & 1 & 2 & 5 & 10 & 20 \\
		
		\midrule
		
		F-Score & 0.728 & 0.732 & \textbf{0.735} & 0.734 & 0.719 \\
		$\text{Precision}_{LTB}$ & 0.728 & 0.736 & \textbf{0.741} & 0.734 & 0.719 \\
		$\text{Recall}_{LTB}$ & 0.728 & 0.728 & 0.729 & \textbf{0.731} & 0.719 \\
		\bottomrule		
\end{tabular}
\end{table}

\begin{figure}[t]
\centering
  \includegraphics[width=.97\linewidth]{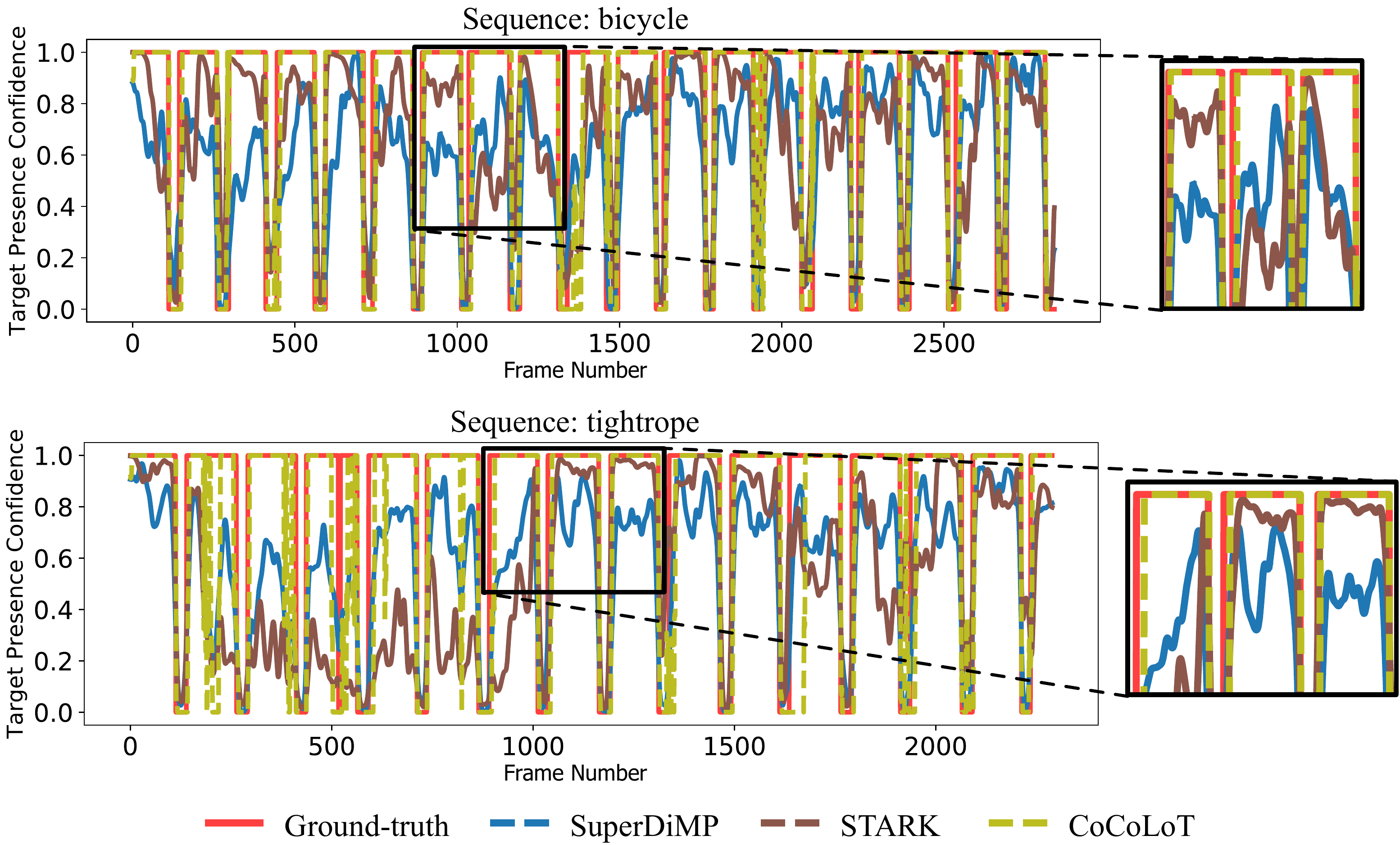}
  \caption{Examples of the ability of the proposed solution in determining the visual presence  of the target (i.e the confidence) in the frames of three different sequences of the LTB-50 \cite{Lukezic2020,VOT2019}. The confidences predicted by STARK and SuperDiMP also reported for comparison. CoCoLoT results in a better and more stable estimation of target presence along the videos.}
  \label{fig:visibilityplots}
\end{figure}

\subsection{Target Presence Determination}
Table \ref{tab:visibilityacc} reports the performance of $\trackerours$ in determining the target presence in the frames of the LTB-50 benchmark \cite{Lukezic2020,VOT2019} in comparison with the underlying trackers. 
Specifically, for each frame in the dataset we compared the target presence label (0 or 1) with each tracker's presence label computed after thresholding at 0.5 the tracker's predicted $\confidence_t^{(i)}$.
The Accuracy reports the average agreement between the ground-truth and tracker-specific labels. The Sensitivity reports the fraction of correctly predicted presences in the frames where the target is actually present. 
The Specificity instead reports the fraction of correct non-presence predictions for those frames in which the target is not present. 
$\trackerours$ presents a higher accuracy (+7.4\%) and sensitivity (+9\%) with respect to the two trackers, meaning that it has a better ability in the determination of the target when this is visible. The Specificity is reduced, suggesting that, despite the overall good ability, the proposed strategy provokes a larger amount of false positives during the absence of the target in the frames. 

Figure \ref{fig:visibilityplots} shows three examples of how the target presence estimation of our proposed methodology better fits with the ground-truth along a video sequence.

Table \ref{tab:spl:visibilityframes} reports the sensibility of the proposed target presence determination strategy in relation to the number of frames $\widehat{T}$.
Employing a number of frames $\widehat{T} > 1$ is preferable but it must be assured that $\widehat{T}$ is not too large (i.e. $\leq 10$) since it could lead to a reduced tracking performance.

\begin{table}[t]
\fontsize{5.5}{6.5}\selectfont
	\centering
	\caption{Performance of CoCoLoT on the LTB-50 benchmark \cite{Lukezic2020,VOT2019} in comparison with the state-of-the-art. Best result, per metric, is highlighted in red, second-best in blue.}
	\label{tab:vot2020sota}
	\setlength\tabcolsep{.04cm}
	\begin{tabular}{l | c c c c c c c c c}
		\toprule
		
		& SPLT & CLGS & DMTrack & LTMU\_B & LT\_DSE & STARK-ST50 & KeepTrack & Zitong et al. & $\trackerours$ \\
		\midrule
		
		F-Score & 0.565 & 0.674 & 0.687 & 0.691 & 0.695 & 0.702 & 0.709 & \tblsecondbest{0.711} & \tblbest{0.735} \\
		$\text{Precision}_{LTB}$ & 0.587 & \tblsecondbest{0.739} & 0.690 & 0.701 & 0.715 & 0.710 & 0.723 & 0.726 & \tblbest{0.741} \\
		$\text{Recall}_{LTB}$ & 0.544 & 0.619 & 0.662 & 0.681 & 0.677 & 0.695 & \tblsecondbest{0.697} & \tblsecondbest{0.697} & \tblbest{0.729} \\
        
		\bottomrule		
\end{tabular}
\end{table}

\subsection{State-of-the-Art Comparison}
In this paragraph, we present the performance of $\trackerours$ in comparison with the state-of-the-art. For all the compared methodologies we report the performance presented in their original papers (the ``-'' symbol reports that the authors did not provide results for the particular measure).

Table \ref{tab:vot2020sota} reports the results of $\trackerours$ against the trackers SPLT \cite{SPLT}, CLGS \cite{VOT2019}, DMTrack \cite{DMTrack}, LTMU\_B \cite{LTMU}, LT\_DSE \cite{LTMU,VOT2019}, STARK-ST50 \cite{Stark}, KeepTrack \cite{KeepTrack}, and the solution of Zitong et al. \cite{Zitong2021}.
Our proposed tracker results the best solution across all the performance measures and hence sets new state-of-the-art results. The improvement over the second-best method \cite{Zitong2021} is of of 3.5\% in F-Score, 2\% in $\text{Precision}_{LTB}$, and of 4.6\% in $\text{Recall}_{LTB}$. 
Our fusion strategy results better than the KeepTrack strategy \cite{KeepTrack},  which augments the long-term capabilities of SuperDiMP by tracking and suppressing distractor objects, and even better than the LTMU\_B pipeline \cite{LTMU} that uses the DiMP tracker in the standard long-term scheme presented in Figure \ref{fig:scheme}(a).

\begin{table}[t]
\fontsize{5.5}{6.5}\selectfont
	\centering
	\caption{Performance achieved by CoCoLoT on the TLP benchmark \cite{TLP} in comparison with the state-of-the-art. Best result, per metric, is highlighted in red, second-best in blue.}
	\label{tab:tlpsota}
	\setlength\tabcolsep{.06cm}
	\begin{tabular}{l | c c c c c c c c }
		\toprule
		
		& SiamFC & MDNet & GlobalTrack & SuperDiMP & DMTrack & STARK-ST50 & LTMU & $\trackerours$ \\
		\midrule

		$\text{Success}_{TLP}$ & 0.237 & 0.365 & 0.520 & 0.537 & 0.541 & 0.549 & \tblsecondbest{0.551} & \tblbest{0.587} \\
		
		$\text{Precision}_{TLP}$ & 0.281 & 0.384 & 0.567  & 0.563 & 0.591 & 0.568 & \tblsecondbest{0.619} & \tblbest{0.633} \\

		\bottomrule		
\end{tabular}
\end{table}

In Table \ref{tab:tlpsota} we compare our solution to SiamFC \cite{SiamFC}, MDNet \cite{MDNet}, GlobalTrack \cite{GlobalTrack}, SuperDiMP, \cite{DiMP,VOT2020} DMTrack \cite{DMTrack}, STARK-ST50 \cite{Stark}, and LTMU \cite{LTMU}
on the TLP benchmark \cite{TLP}. $\trackerours$ results again the best tracker over both the considered metrics. Particularly, the $\text{Success}_{TLP}$ improvement of the underlying trackers STARK-ST50 and SuperDiMP is of 6.9\% and 9.3\% respectively.

\begin{table}[t]
\fontsize{5.5}{6.5}\selectfont
	\centering
	\caption{Performance of CoCoLoT on the LaSOT benchmark \cite{LaSOT} in comparison with the state-of-the-art. Best result, per metric, is highlighted in red, second-best in blue.}
	\label{tab:lasotsota}
	\setlength\tabcolsep{.05cm}
	\begin{tabular}{l | c c c c c c c c}
		\toprule
		& SuperDiMP & TrDiMP & LTMU & Siam-R-CNN & TransT & STARK-ST50 & KeepTrack  & $\trackerours$ \\
		 
		\midrule

		$\text{Success}_{LaSOT}$ & 0.631 & 0.639 & 0.647 & 0.648 & 0.649 & 0.664 & \tblsecondbest{0.671} & \tblbest{0.685} \\
		$\text{Precision}_{LaSOT}$ & 0.653 & 0.663 & 0.665 & 0.684 & 0.690 & 0.693 & \tblsecondbest{0.702} & \tblbest{0.725} \\
        
		\bottomrule		
\end{tabular}
\end{table}

The LaSOT benchmark \cite{LaSOT} comparison (presented in Table \ref{tab:lasotsota}) of $\trackerours$ with the trackers SuperDiMP \cite{DiMP,VOT2020}, TrDiMP \cite{TrDiMP2}, LTMU \cite{LTMU}, Siam-R-CNN \cite{SiamRCNN}, TransT \cite{TransT}, STARK-ST50 \cite{Stark}, and KeepTrack \cite{KeepTrack}, additionally confirms the effectiveness of the proposed solution for long-term visual object tracking scenarios.

\section{Conclusions}
In this paper, we focused on the problem of fusing the capabilities of complementary visual trackers in long-term scenarios. 
We proposed CoCoLoT, a framework in which an effective evaluation strategy based on an online learned deep learning model is used to assess the behavior of the underlying trackers STARK \cite{Stark} and SuperDiMP \cite{DiMP,VOT2020}. Based on the proposed evaluation function, a selection policy was implemented to select which of the two trackers provides the best target localization. Such an outcome was used to localize the target and also to correct the performance of the non-selected tracker, ultimately correcting both trackers from errors by themselves. We provided extensive experimental results to understand the impact of the modules composing our solution. Results are 
very interesting and 
competitive with the state-of-the-art on the LTB-50 \cite{Lukezic2020,VOT2019}, TLP \cite{TLP}, and LaSOT \cite{LaSOT} benchmarks.

\small\textbf{Ack.} Research supported by the ACHIEVE-ITN H2020 project.

\bibliographystyle{IEEEtran}
\bibliography{IEEEabrv,egbib}

\begin{thebibliography}{10}
\providecommand{\url}[1]{#1}
\csname url@samestyle\endcsname
\providecommand{\newblock}{\relax}
\providecommand{\bibinfo}[2]{#2}
\providecommand{\BIBentrySTDinterwordspacing}{\spaceskip=0pt\relax}
\providecommand{\BIBentryALTinterwordstretchfactor}{4}
\providecommand{\BIBentryALTinterwordspacing}{\spaceskip=\fontdimen2\font plus
\BIBentryALTinterwordstretchfactor\fontdimen3\font minus
  \fontdimen4\font\relax}
\providecommand{\BIBforeignlanguage}[2]{{%
\expandafter\ifx\csname l@#1\endcsname\relax
\typeout{** WARNING: IEEEtran.bst: No hyphenation pattern has been}%
\typeout{** loaded for the language `#1'. Using the pattern for}%
\typeout{** the default language instead.}%
\else
\language=\csname l@#1\endcsname
\fi
#2}}
\providecommand{\BIBdecl}{\relax}
\BIBdecl

\bibitem{micheloni2010intelligent}
C.~Micheloni, P.~Remagnino, H.-L. Eng, and J.~Geng, ``Intelligent monitoring of
  complex environments,'' \emph{IEEE Intelligent Systems}, vol.~25, no.~3, pp.
  12--14, 2010.

\bibitem{Dunnhofer2021trek150}
M.~Dunnhofer, A.~Furnari, G.~M. Farinella, and C.~Micheloni, ``Is first person
  vision challenging for object tracking?'' in \emph{Proceedings of the
  IEEE/CVF International Conference on Computer Vision Workshops}, 2021, pp.
  2698--2710.

\bibitem{RE3}
D.~Gordon, A.~Farhadi, and D.~Fox, ``{Re 3 : Real-time recurrent regression
  networks for visual tracking of generic objects},'' \emph{IEEE Robotics and
  Automation Letters}, vol.~3, no.~2, pp. 788--795, 2018.

\bibitem{Dunnhofer2020MedIA}
M.~Dunnhofer, M.~Antico, F.~Sasazawa, Y.~Takeda, S.~Camps, N.~Martinel,
  C.~Micheloni, G.~Carneiro, and D.~Fontanarosa, ``{Siam-U-Net: encoder-decoder
  siamese network for knee cartilage tracking in ultrasound images},''
  \emph{Medical Image Analysis}, vol.~60, p. 101631, feb 2020.

\bibitem{Lukezic2020}
A.~Lukeźič, L.~C. Zajc, T.~Vojíř, J.~Matas, and M.~Kristan, ``Performance
  evaluation methodology for long-term single-object tracking,'' \emph{IEEE
  Transactions on Cybernetics}, pp. 1--14, 2020.

\bibitem{OTB}
Y.~Wu, J.~Lim, and M.~H. Yang, ``{Object tracking benchmark},'' \emph{IEEE
  Transactions on Pattern Analysis and Machine Intelligence}, vol.~37, no.~9,
  pp. 1834--1848, sep 2015.

\bibitem{VOT2019}
M.~Kristan, J.~Matas, A.~Leonardis, M.~Felsberg, R.~Pflugfelder, J.-K.
  K{\"{a}}m{\"{a}}r{\"{a}}inen, L.~Zajc, O.~Drbohlav, A.~Luke{\v{z}}i{\v{c}},
  A.~Berg, A.~Eldesokey, J.~K{\"{a}}pyl{\"{a}}, G.~Fern{\'{a}}ndez
  \emph{et~al.}, ``{The Seventh Visual Object Tracking VOT2019 Challenge
  Results},'' in \emph{Proceedings of the IEEE/CVF International Conference on
  Computer Vision Workshops}, 2019.

\bibitem{UAV123}
\BIBentryALTinterwordspacing
M.~Mueller, N.~Smith, and B.~Ghanem, ``{A Benchmark and Simulator for UAV
  Tracking},'' in \emph{European Conference on Computer Vision}.\hskip 1em plus
  0.5em minus 0.4em\relax Springer, Cham, 2016, pp. 445--461. [Online].
  Available: \url{http://link.springer.com/10.1007/978-3-319-46448-0{\_}27}
\BIBentrySTDinterwordspacing

\bibitem{TC128}
P.~Liang, E.~Blasch, and H.~Ling, ``{Encoding Color Information for Visual
  Tracking: Algorithms and Benchmark},'' \emph{IEEE Transactions on Image
  Processing}, vol.~24, no.~12, pp. 5630--5644, dec 2015.

\bibitem{NfS}
\BIBentryALTinterwordspacing
H.~K. Galoogahi, A.~Fagg, C.~Huang, D.~Ramanan, and S.~Lucey, ``{Need for
  Speed: A Benchmark for Higher Frame Rate Object Tracking},'' in
  \emph{Proceedings of the IEEE International Conference on Computer Vision},
  vol. 2017-Octob.\hskip 1em plus 0.5em minus 0.4em\relax Institute of
  Electrical and Electronics Engineers Inc., mar 2017, pp. 1134--1143.
  [Online]. Available: \url{http://arxiv.org/abs/1703.05884}
\BIBentrySTDinterwordspacing

\bibitem{ECO}
\BIBentryALTinterwordspacing
M.~Danelljan, G.~Bhat, F.~S. Khan, and M.~Felsberg, ``{ECO: Efficient
  Convolution Operators for Tracking},'' in \emph{IEEE Conference on Computer
  Vision and Pattern Recognition}, nov 2017. [Online]. Available:
  \url{http://arxiv.org/abs/1611.09224}
\BIBentrySTDinterwordspacing

\bibitem{ATOM}
\BIBentryALTinterwordspacing
------, ``{ATOM: Accurate Tracking by Overlap Maximization},'' in \emph{IEEE
  Conference on Computer Vision and Pattern Recognition}, 2019. [Online].
  Available: \url{https://github.com/visionml/pytracking.
  http://arxiv.org/abs/1811.07628}
\BIBentrySTDinterwordspacing

\bibitem{DiMP}
\BIBentryALTinterwordspacing
G.~Bhat, M.~Danelljan, L.~{Van Gool}, and R.~Timofte, ``{Learning
  Discriminative Model Prediction for Tracking},'' in \emph{Proceedings of the
  IEEE/CVF International Conference on Computer Vision}, 2019. [Online].
  Available: \url{https://github.com/visionml/pytracking.
  http://arxiv.org/abs/1904.07220}
\BIBentrySTDinterwordspacing

\bibitem{SiamFC}
\BIBentryALTinterwordspacing
L.~Bertinetto, J.~Valmadre, J.~F. Henriques, A.~Vedaldi, and P.~H. Torr,
  ``{Fully-convolutional siamese networks for object tracking},''
  \emph{European Conference on Computer Vision}, vol. 9914 LNCS, pp. 850--865,
  2016. [Online]. Available: \url{http://arxiv.org/abs/1606.09549}
\BIBentrySTDinterwordspacing

\bibitem{SiamRPNpp}
\BIBentryALTinterwordspacing
B.~Li, W.~Wu, Q.~Wang, F.~Zhang, J.~Xing, and J.~Yan, ``{SIAMRPN++: Evolution
  of siamese visual tracking with very deep networks},'' in \emph{Proceedings
  of the IEEE Computer Society Conference on Computer Vision and Pattern
  Recognition}, vol. 2019-June, 2019, pp. 4277--4286. [Online]. Available:
  \url{http://bo-li.info/SiamRPN++.}
\BIBentrySTDinterwordspacing

\bibitem{SiamGAT}
D.~Guo, Y.~Shao, Y.~Cui, Z.~Wang, L.~Zhang, and C.~Shen, ``Graph attention
  tracking,'' in \emph{Proceedings of the IEEE/CVF Conference on Computer
  Vision and Pattern Recognition (CVPR)}, June 2021, pp. 9543--9552.

\bibitem{GOTURN}
\BIBentryALTinterwordspacing
D.~Held, S.~Thrun, and S.~Savarese, ``{Learning to Track at 100 FPS with Deep
  Regression Networks},'' \emph{European Conference on Computer Vision}, vol.
  abs/1604.0, 2016. [Online]. Available: \url{http://arxiv.org/abs/1604.01802}
\BIBentrySTDinterwordspacing

\bibitem{Dunnhofer2019}
\BIBentryALTinterwordspacing
M.~Dunnhofer, N.~Martinel, G.~L. Foresti, and C.~Micheloni, ``{Visual Tracking
  by means of Deep Reinforcement Learning and an Expert Demonstrator},'' in
  \emph{Proceedings of The IEEE/CVF International Conference on Computer Vision
  Workshops}, 2019. [Online]. Available: \url{http://arxiv.org/abs/1909.08487}
\BIBentrySTDinterwordspacing

\bibitem{Dunnhofer2021ral}
M.~Dunnhofer, N.~Martinel, and C.~Micheloni, ``Weakly-supervised domain
  adaptation of deep regression trackers via reinforced knowledge
  distillation,'' \emph{IEEE Robotics and Automation Letters}, vol.~6, no.~3,
  pp. 5016--5023, 2021.

\bibitem{TrDiMP2}
N.~Wang, W.~Zhou, J.~Wang, and H.~Li, ``Transformer meets tracker exploiting
  temporal context for robust visual tracking,'' in \emph{Proceedings of the
  IEEE/CVF Conference on Computer Vision and Pattern Recognition (CVPR)}, June
  2021, pp. 1571--1580.

\bibitem{TransT}
X.~Chen, B.~Yan, J.~Zhu, D.~Wang, X.~Yang, and H.~Lu, ``Transformer tracking,''
  in \emph{Proceedings of the IEEE/CVF Conference on Computer Vision and
  Pattern Recognition (CVPR)}, June 2021, pp. 8126--8135.

\bibitem{Stark}
\BIBentryALTinterwordspacing
B.~Yan, H.~Peng, J.~Fu, D.~Wang, and H.~Lu, ``Learning spatio-temporal
  transformer for visual tracking,'' in \emph{Proceedings of the IEEE/CVF
  International Conference on Computer Vision (ICCV}, 2021. [Online].
  Available: \url{https://arxiv.org/abs/2103.17154}
\BIBentrySTDinterwordspacing

\bibitem{TLD}
Z.~Kalal, K.~Mikolajczyk, and J.~Matas, ``{Tracking-learning-detection},''
  \emph{IEEE Transactions on Pattern Analysis and Machine Intelligence},
  vol.~34, no.~7, pp. 1409--1422, 2012.

\bibitem{PTAV}
H.~Fan and H.~Ling, ``Parallel tracking and verifying,'' \emph{IEEE
  Transactions on Image Processing}, vol.~28, no.~8, pp. 4130--4144, 2019.

\bibitem{SPLT}
\BIBentryALTinterwordspacing
B.~Yan, H.~Zhao, D.~Wang, H.~Lu, and X.~Yang, ``{'Skimming-perusal' tracking: A
  framework for real-time and robust long-term tracking},'' in
  \emph{Proceedings of the IEEE International Conference on Computer Vision},
  vol. 2019-Octob, 2019, pp. 2385--2393. [Online]. Available:
  \url{https://github.com/iiau-tracker/SPLT.}
\BIBentrySTDinterwordspacing

\bibitem{LTMU}
K.~Dai, Y.~Zhang, D.~Wang, J.~Li, H.~Lu, and X.~Yang, ``{High-Performance
  Long-Term Tracking With Meta-Updater},'' in \emph{IEEE/CVF Conference on
  Computer Vision and Pattern Recognition}.\hskip 1em plus 0.5em minus
  0.4em\relax Institute of Electrical and Electronics Engineers (IEEE), aug
  2020, pp. 6297--6306.

\bibitem{VOT2020}
M.~Kristan, A.~Leonardis, J.~Matas, M.~Felsberg, R.~Pflugfelder, J.-K.
  K{\"a}m{\"a}r{\"a}inen, M.~Danelljan, L.~{\v{C}}. Zajc,
  A.~Luke{\v{z}}i{\v{c}}, O.~Drbohlav, L.~He, Y.~Zhang, S.~Yan, J.~Yang,
  G.~Fern{\'a}ndez \emph{et~al.}, ``The eighth visual object tracking vot2020
  challenge results,'' in \emph{Computer Vision -- ECCV 2020 Workshops},
  A.~Bartoli and A.~Fusiello, Eds.\hskip 1em plus 0.5em minus 0.4em\relax Cham:
  Springer International Publishing, 2020, pp. 547--601.

\bibitem{Yoon2012}
J.~H. Yoon, D.~Y. Kim, and K.~J. Yoon, ``{Visual tracking via adaptive tracker
  selection with multiple features},'' in \emph{European Conference on Computer
  Vision}, vol. 7575 LNCS, no. PART 4, 2012, pp. 28--41.

\bibitem{Bailer2014}
C.~Bailer, A.~Pagani, and D.~Stricker, ``{A superior tracking approach:
  Building a strong tracker through fusion},'' in \emph{European Conference on
  Computer Vision}, vol. 8695 LNCS, no. PART 7.\hskip 1em plus 0.5em minus
  0.4em\relax Springer Verlag, 2014, pp. 170--185.

\bibitem{Vojir2016}
T.~Vojir, J.~Matas, and J.~Noskova, ``{Online adaptive hidden Markov model for
  multi-tracker fusion},'' \emph{Computer Vision and Image Understanding}, vol.
  153, pp. 109--119, dec 2016.

\bibitem{Dunnhofer2020accv}
M.~Dunnhofer, N.~Martinel, and C.~Micheloni, ``{Tracking-by-Trackers with a
  Distilled and Reinforced Model},'' in \emph{Asian Conference on Computer
  Vision}, 2020.

\bibitem{TLP}
A.~Moudgil and V.~Gandhi, ``Long-term visual object tracking benchmark,'' in
  \emph{Asian Conference on Computer Vision}.\hskip 1em plus 0.5em minus
  0.4em\relax Springer, 2018, pp. 629--645.

\bibitem{LaSOT}
\BIBentryALTinterwordspacing
H.~Fan, L.~Lin, F.~Yang, P.~Chu, G.~Deng, S.~Yu, H.~Bai, Y.~Xu, C.~Liao, and
  H.~Ling, ``{LaSOT: A High-quality Benchmark for Large-scale Single Object
  Tracking},'' in \emph{IEEE Conference on Computer Vision and Pattern
  Recognition}, sep 2019. [Online]. Available:
  \url{http://arxiv.org/abs/1809.07845}
\BIBentrySTDinterwordspacing

\bibitem{Fucolot}
A.~Luke{\v{z}}i{\v{c}}, L.~{\v{C}}. Zajc, T.~Voj{\'i}{\v{r}}, J.~Matas, and
  M.~Kristan, ``Fucolot -- a fully-correlational long-term tracker,'' in
  \emph{Computer Vision -- ACCV 2018}, C.~V. Jawahar, H.~Li, G.~Mori, and
  K.~Schindler, Eds.\hskip 1em plus 0.5em minus 0.4em\relax Cham: Springer
  International Publishing, 2019, pp. 595--611.

\bibitem{MOSSE}
D.~S. Bolme, J.~R. Beveridge, B.~A. Draper, and Y.~M. Lui, ``{Visual object
  tracking using adaptive correlation filters},'' in \emph{IEEE Conference on
  Computer Vision and Pattern Recognition}.\hskip 1em plus 0.5em minus
  0.4em\relax IEEE, 2010, pp. 2544--2550.

\bibitem{GlobalTrack}
\BIBentryALTinterwordspacing
L.~Huang, X.~Zhao, and K.~Huang, ``{GlobalTrack: A Simple and Strong Baseline
  for Long-term Tracking},'' in \emph{AAAI Conference on Artificial
  Intelligence}, dec 2020. [Online]. Available:
  \url{http://arxiv.org/abs/1912.08531}
\BIBentrySTDinterwordspacing

\bibitem{Attention}
\BIBentryALTinterwordspacing
A.~Vaswani, N.~Shazeer, N.~Parmar, J.~Uszkoreit, L.~Jones, A.~N. Gomez, L.~u.
  Kaiser, and I.~Polosukhin, ``Attention is all you need,'' in \emph{Advances
  in Neural Information Processing Systems}, I.~Guyon, U.~V. Luxburg,
  S.~Bengio, H.~Wallach, R.~Fergus, S.~Vishwanathan, and R.~Garnett, Eds.,
  vol.~30.\hskip 1em plus 0.5em minus 0.4em\relax Curran Associates, Inc.,
  2017. [Online]. Available:
  \url{https://proceedings.neurips.cc/paper/2017/file/3f5ee243547dee91fbd053c1c4a845aa-Paper.pdf}
\BIBentrySTDinterwordspacing

\bibitem{MEEM}
J.~Zhang, S.~Ma, and S.~Sclaroff, ``{MEEM: Robust tracking via multiple experts
  using entropy minimization},'' in \emph{European Conference on Computer
  Vision}, vol. 8694 LNCS, no. PART 6.\hskip 1em plus 0.5em minus 0.4em\relax
  Springer Verlag, 2014, pp. 188--203.

\bibitem{Wang2014}
N.~Wang and D.~Y. Yeung, ``{Ensemble-based tracking: Aggregating crowdsourced
  structured time series data},'' in \emph{31st International Conference on
  Machine Learning, ICML 2014}, vol.~4, 2014, pp. 2807--2817.

\bibitem{Song2020}
\BIBentryALTinterwordspacing
k.~Song, W.~Zhang, R.~Song, and Y.~Li, ``Online decision based visual tracking
  via reinforcement learning,'' in \emph{Advances in Neural Information
  Processing Systems}, H.~Larochelle, M.~Ranzato, R.~Hadsell, M.~F. Balcan, and
  H.~Lin, Eds., vol.~33.\hskip 1em plus 0.5em minus 0.4em\relax Curran
  Associates, Inc., 2020, pp. 11\,778--11\,788. [Online]. Available:
  \url{https://proceedings.neurips.cc/paper/2020/file/885b2c7a6deb4fea10f319c4ce993e02-Paper.pdf}
\BIBentrySTDinterwordspacing

\bibitem{RLTDiMP}
\BIBentryALTinterwordspacing
S.~Choi, J.~Lee, Y.~Lee, and A.~G. Hauptmann, ``Robust long-term object
  tracking via improved discriminative model prediction,'' vol. abs/2008.04722,
  2020. [Online]. Available: \url{https://arxiv.org/abs/2008.04722}
\BIBentrySTDinterwordspacing

\bibitem{MDNet}
H.~Nam and B.~Han, ``{Learning Multi-domain Convolutional Neural Networks for
  Visual Tracking},'' \emph{IEEE Conference on Computer Vision and Pattern
  Recognition}, vol. 2016-Decem, pp. 4293--4302, 2016.

\bibitem{VOT2021}
M.~Kristan, J.~Matas, A.~Leonardis, M.~Felsberg, R.~Pflugfelder, J.-K.
  K{\"a}m{\"a}r{\"a}inen, H.~J. Chang, M.~Danelljan, L.~Cehovin,
  A.~Luke{\v{z}}i{\v{c}} \emph{et~al.}, ``The ninth visual object tracking
  vot2021 challenge results,'' in \emph{Proceedings of the IEEE/CVF
  International Conference on Computer Vision}, 2021, pp. 2711--2738.

\bibitem{DMTrack}
Z.~Zhang, B.~Zhong, S.~Zhang, Z.~Tang, X.~Liu, and Z.~Zhang, ``Distractor-aware
  fast tracking via dynamic convolutions and mot philosophy,'' in
  \emph{Proceedings of the IEEE/CVF Conference on Computer Vision and Pattern
  Recognition}, 2021, pp. 1024--1033.

\bibitem{KeepTrack}
C.~Mayer, M.~Danelljan, D.~P. Paudel, and L.~Van~Gool, ``Learning target
  candidate association to keep track of what not to track,'' in
  \emph{Proceedings of the IEEE/CVF International Conference on Computer Vision
  (ICCV)}, October 2021.

\bibitem{Zitong2021}
Z.~Yi, Z.~Tong, Y.~Zhao, Z.~Zhao, and F.~Su, ``A method of stable long-term
  single object tracking,'' in \emph{2021 IEEE International Conference on
  Multimedia and Expo (ICME)}, 2021, pp. 1--6.

\bibitem{SiamRCNN}
P.~Voigtlaender, J.~Luiten, P.~H. Torr, and B.~Leibe, ``Siam r-cnn: Visual
  tracking by re-detection,'' in \emph{IEEE/CVF Conference on Computer Vision
  and Pattern Recognition (CVPR)}, June 2020.

\end{thebibliography}

\end{document}